# Solving Weighted Constraint Satisfaction Problems with Memetic/Exact Hybrid Algorithms

**José E. Gallardo**　　　　　　　　　　　　　　　　　　PEPEG@LCC.UMA.ES
**Carlos Cotta**　　　　　　　　　　　　　　　　　　CCOTTAP@LCC.UMA.ES
**Antonio J. Fernández**　　　　　　　　　　　　　　　　AFDEZ@LCC.UMA.ES
*Dept. Lenguajes y Ciencias de la Computación*
*Universidad de Málaga, ETSI Informática*
*Campus de Teatinos, 29071 – Málaga, Spain*

## Abstract

A *weighted constraint satisfaction problem* (WCSP) is a constraint satisfaction problem in which preferences among solutions can be expressed. Bucket elimination is a complete technique commonly used to solve this kind of constraint satisfaction problem. When the memory required to apply bucket elimination is too high, a heuristic method based on it (denominated mini-buckets) can be used to calculate bounds for the optimal solution. Nevertheless, the curse of dimensionality makes these techniques impractical on large scale problems. In response to this situation, we present a memetic algorithm for WCSPs in which bucket elimination is used as a mechanism for recombining solutions, providing the best possible child from the parental set. Subsequently, a multi-level model in which this exact/metaheuristic hybrid is further hybridized with branch-and-bound techniques and mini-buckets is studied. As a case study, we have applied these algorithms to the resolution of the maximum density still life problem, a hard constraint optimization problem based on Conway's game of life. The resulting algorithm consistently finds optimal patterns for up to date solved instances in less time than current approaches. Moreover, it is shown that this proposal provides new best known solutions for very large instances.

## 1. Introduction

Many real problems can be formulated as constraint satisfaction problems (CSPs) in which solutions are assignments to a set of variables (each variable taking values from a certain domain), and in which there exists a collection of constraints that restrict the assignment of particular values or combination of values; solving a CSP means finding a feasible assignment of values to variables, i.e., one where all the constraints are satisfied. However, a wide range of problems cannot be posed this way, either because the problem is over-constrained (and thus there is no solution) or because the problem has multiple solutions and the objective is to find the best one according to some optimality criterion. In both cases, the problem might be handled from an optimization point of view by associating preferences to the constraints. This kind of CSP in which preferences among constraints/solutions can be expressed are called *weighted constraint satisfaction problems* (WCSPs) (Schiex, Fargier, & Verfaillie, 1995; Bistarelli, Montanari, & Rossi, 1997). Solving a WCSP means optimally satisfying a set of weighted constraints. This clearly enlarges the scope of CSPs: many practical problems can be modeled as WCSPs, such as for instance, radio frequency assignment, scheduling and cellular manufacturing, among others (Cabon, de Givry, Lobjois, Schiex, & Warners, 1999; Khemmoudj & Bennaceur, 2007; Nonobe & Ibaraki, 2001).

Complete methods, like branch-and-bound (Lawler & Wood, 1966) and bucket elimination (Dechter, 1999), a technique which originated in the early work of Bertele and Brioschi (1972) on nonserial dynamic programming, are two of the most popular ways to attack WCSPs. However, although the picture of a CSP is very general, it should be noted that the inclusion of preferences in its constraints makes a particular WCSP very specific and as a consequence WCSPs have to be tackled using very specialized algorithms that were specifically designed (Freuder & Wallace, 1992; Verfaillie, Lemaître, & Schiex, 1996; Kask & Detcher, 2001; Lemaître, Verfaillie, Bourreau, & Laburthe, 2001; Larrosa & Schiex, 2004; Gelain, Pini, Rossi, & Venable, 2007; Khemmoudj & Bennaceur, 2007; Marinescu & Dechter, 2007). Moreover, general techniques require a very large computational effort (in time, memory or both) to solve many WCSPs, due to their size and complexity, and therefore are impractical in many cases. This can be alleviated using heuristic methods, e.g., beam search (BS) (Barr & Feigenbaum, 1981) and mini-buckets (Dechter, 1997), for branch-and-bound and bucket elimination respectively. However, in large scale problems, the high computational cost is still evident.

In this context the use of alternative techniques must be considered to overcome the limitations of general techniques; for instance, evolutionary algorithms (Bäck, 1996; Bäck, Fogel, & Michalewicz, 1997) are powerful heuristics for optimization problems based on the principles of natural evolution, which are flexible enough to be deployed in a wide range of problems. However, this *generality* reduces their competitiveness, unless domain knowledge is also incorporated. This need for exploiting domain knowledge in optimization methods has been repeatedly shown (Wolpert & Macready, 1997; Culberson, 1998), and memetic algorithms (Moscato & Cotta, 2003, 2007; Krasnogor & Smith, 2005) represent one of the most successful responses to this need (Hart, Krasnogor, & Smith, 2005). This paper explores different ways of hybridizing branch-and-bound/bucket elimination (and their corresponding heuristic methods) and memetic algorithms, combining their search capabilities in a synergetic way.

The hybrid techniques proposed here can be used as general problem solvers for WCSPs. Note that they are essentially heuristic in nature and hence they cannot provide optimality proofs for the solutions they obtain. Notice however that they can probably provide optimal or near-optimal solutions to a wide range of WCSPs. Furthermore, these hybrid techniques are less time-consuming than the general methods involved in them, and can thus be applied to larger problem instances. In order to experimentally evaluate the hybrid techniques, we have tackled the Maximum Density Still Life Problem, a very hard combinatorial optimization problem which is also a prime example of a weighted constraint optimization problem. No polynomial-time algorithm is known to solve this problem, although, to the best of our knowledge, the problem has not yet been proven to be NP-hard. For these reasons, it is not surprising that this problem has attracted the interest of the constraint-programming community, and has been central in the development and assessment of sophisticated techniques such as bucket elimination. Indeed, it constitutes an excellent test bed for different optimization techniques, and has been included in the CSPLib[1] repository. A web page[2] keeps record of up-to-date results.

---

1. http://www.csplib.org
2. http://www.ai.sri.com/~nysmith/life

## 2. Preliminaries

In this section, we briefly introduce concepts and techniques that will be used in the rest of the paper. To this end, we first define weighted constraint satisfaction problems, as well as the techniques of bucket elimination and mini-buckets. Subsequently, we describe beam search, a heuristic tree search algorithm derived from branch-and-bound. Finally, memetic algorithms are presented. For the sake of notational simplicity, where appropriate we stick to the notation of Larrosa et al. (2003, 2005).

### 2.1 Weighted Constraint Satisfaction Problems

A *weighted constraint satisfaction problem* (WCSP) (Schiex et al., 1995; Bistarelli et al., 1997) is a constraint satisfaction problem (CSP) in which preferences among solutions can be expressed. Formally, a WCSP can be defined by a tuple $(\mathcal{X}, \mathcal{D}, \mathcal{F})$, where $\mathcal{D} = \{D_1, \cdots, D_n\}$ is a set of *finite domains*, $\mathcal{X} = \{x_1, \cdots, x_n\}$ is a set of variables taking values from their finite domains ($D_i$ is the domain of variable $x_i$) and $\mathcal{F}$ is a set of *cost functions* (also called *soft constraints* or *weighted constraints*) used to declare preferences among possible solutions. Variable correctly assigned receive finite costs that express their degree of preference (the lower the value the better the preference) and variables not correctly assigned receive cost $\infty$. Note that each $f \in \mathcal{F}$ is defined over a subset of variables, $var(f) \subseteq \mathcal{X}$, called its *scope*. The objective function $F$ is defined as the sum of all functions in $\mathcal{F}$, i.e., $F = \sum_{f \in \mathcal{F}} f$.

The assignment of value $v_i \in D_i$ to variable $x_i$ is noted $x_i = v_i$. A partial assignment of $m < n$ variables is a tuple $t = (x_{i_1} = v_1, x_{i_2} = v_2, \cdots, x_{i_m} = v_m)$ where $i_j \in \{1, \ldots, n\}$ are all different. A complete assignment of all variables with values in their domains that satisfies every soft constraint (i.e., with a finite valuation for $F$) represents a solution to the WCSP. The optimization goal is to find a solution that minimizes this objective function.

### 2.2 Bucket Elimination

Bucket elimination (BE) (Dechter, 1999) is a generic technique suitable for many automated reasoning and optimization problems and, in particular, for solving WCSP. The functioning of BE is based upon the following two operators over functions (Larrosa et al., 2005):

- the sum of two functions $f$ and $g$, denoted $(f + g)$, is a new function with scope $var(f) \cup var(g)$ which returns for each tuple the sum of costs of $f$ and $g$, i.e., $(f+g)(t) = f(t) + g(t)$.

- The elimination of variable $x_i$ from $f$, denoted $f \Downarrow x_i$, is a new function with scope $var(f) - \{x_i\}$ which returns for each tuple $t$ the minimum cost extension of $t$ to $x_i$, $(f \Downarrow x_i)(t) = min_{v \in D_i}\{f(t \cdot (x_i = v))\}$, where $t \cdot (x_i = v)$ means the extension of the assignment $t$ with the assignment of value $v$ to variable $x_i$. Observe that when $f$ is a unary function (i.e., it has arity one), a constant is obtained upon elimination of the only variable in its scope.

Without losing of generality, let us assume a lexicographic ordering for the variables in $\mathcal{X}$, i.e., $o = (x_1, x_2, \cdots, x_n)$. Figure 1 shows pseudo-code of the BE algorithm for solving a WCSP instance, which returns the optimal cost in $\mathcal{F}$ and one optimal assignment in

---

**Bucket Elimination for a WCSP $(\mathcal{X}, \mathcal{D}, F)$**

---

      **function** BE$(\mathcal{X}, \mathcal{D}, \mathcal{F})$
1:    **for** $i := n$ **downto** $1$ **do**
2:       $B_i := \{f \in \mathcal{F} \mid x_i \in var(f)\}$
3:       $g_i := (\sum_{f \in B_i} f) \Downarrow x_i$
4:       $\mathcal{F} := (\mathcal{F} \bigcup \{g_i\}) - B_i$
5:    **end for**
6:    $t := \emptyset$
7:    **for** $i := 1$ **to** $n$ **do**
8:       $v := argmin_{a \in D_i}\{(\sum_{f \in B_i} f)(t \cdot (x_i = a))\}$
9:       $t := t \cdot (x_i = v)$
10:   **end for**
11:   **return**$(\mathcal{F}, t)$
      **end function**

---

Figure 1: The general template, adapted from Larrosa and Morancho (2003), of bucket elimination for a WCSP $(\mathcal{X}, \mathcal{D}, F)$.

$t$. Observe that, initially, BE eliminates in decreasing order one variable $x_i \in \mathcal{X}$ in each iteration of the loop comprising lines 1-5. This is done by computing firstly the bucket $B_i$ of variable $x_i$ as the set of all cost functions in $\mathcal{F}$ having $x_i$ in their scope. Then, a new function $g_i$ is defined as the sum of all these functions in $B_i$ in which variable $x_i$ has been eliminated. Finally, $\mathcal{F}$ is updated by removing the functions involving $x_i$ (i.e., those in $B_i$) and adding the new function that does not contain $x_i$. The consequence is that $x_i$ does not exist in $\mathcal{F}$ but the value of the optimal cost is preserved. The elimination of $x_1$ produces a function with an empty scope (i.e., a constant) which is the optimal cost of the problem. Then, in lines 6-10, BE generates an optimal assignment of variables by considering these in the order imposed by $o$: this is done by starting from an empty assignment $t$ and assigning to $x_i$ the best value of the extension of $t$, with respect to the sum of functions in $B_i$ ($argmin_a\{f(a)\}$ represents the value of $a$ producing the minimum $f(a)$).

Note that BE has exponential space complexity because, in general, the result of summing functions or eliminating variables cannot be expressed intensionally by algebraic expressions and, as a consequence, intermediate results have to be collected extensionally in tables. To be precise, the complexity of BE depends on the problem structure (as captured by its constraint graph $G$) and the ordering $o$. According to Larrosa and Morancho (2003), the complexity of BE along ordering $o$ is time $\Theta(Q \times n \times d^{w^*(o)+1})$ and space $\Theta(n \times d^{w^*(o)})$, where $d$ is the largest domain size, $Q$ is the cost of evaluating cost functions (usually assumed $\Theta(1)$), and $w^*(o)$ is the *induced width* of the graph along ordering $o$, which describes the largest clique created in the graph by bucket elimination, and which corresponds to the largest scope of a function recorded by the algorithm. Although finding the optimal ordering is NP-hard (Arnborg, 1985), heuristics and approximation algorithms have been developed for this task – check the work of Dechter (1999) for details.

### 2.3 Mini-Buckets

The main drawback of BE is that it requires exponential space to store functions extensionally. When this complexity is too high, the solution can be approximated using the mini-bucket (MB) approach presented by Dechter (1997) and Detcher and Rish (2003). Recall that, in order to eliminate variable $x_i$, with its corresponding bucket $B_i = \{f_{i_1}, \ldots, f_{i_m}\}$, BE calculates a new cost function $g_i = (\sum_{f \in B_i} f) \Downarrow x_i$, whose time and space complexity increases with the cardinality of $g_i$, i.e., with the size of the set $\cup_{f \in B_i} var(f) - \{x_i\}$. This complexity can be decreased by approximating the function $g_i$ with a set of smaller-arity functions. The basic idea is to partition bucket $B_i$ into $k$ so called mini-buckets $B_{i_1}, \ldots, B_{i_k}$, such that the number of variables in the scope of each $B_{i_j}$ is bounded by a parameter. Afterwards, a set of $k$ cost functions with the reduced arity sought can be defined as $g_{i_j} = (\sum_{f \in B_{i_j}} f) \Downarrow x_i, j = 1 \ldots k$, and the required approximation to $g_i$ can be computed as the sum $g_i' = \sum_{1 \leqslant j \leqslant k} g_{i_j} = \sum_{1 \leqslant j \leqslant k} ((\sum_{f \in B_{i_j}} f) \Downarrow x_i)$.

Note that the minimization computed in $g_i$ by the $\Downarrow$ operator has been migrated inside the sum. Since, in general, for any two non-negative functions $f_1(x)$ and $f_2(x)$, $min_x(f_1(x) + f_2(x)) \geqslant min_x f_1(x) + min_x f_2(x)$, it follows that $g_i'$ is a lower bound on $g_i$. Therefore, if variable elimination is performed using approximated cost functions, it provides a lower bound for the optimal cost requiring less computation than BE. Notice that the described approach provides a family of under-estimating heuristic functions whose complexity and accuracy is parameterized by the maximum number of variables allowed in each mini-bucket.

### 2.4 Beam Search

Branch-and-bound (BB) (Lawler & Wood, 1966) is a general tree search method for solving combinatorial optimization problems. Tree search methods are constructive, in the sense that they work on partial solutions. In this way, tree search methods start with an empty solution that is incrementally extended by adding components to it. The way that partial solutions can be extended depends on the constraints imposed by the problem being solved. The solution construction mechanism maps the search space to a tree structure, in such a way that a path from the root of the tree to a leaf node corresponds to the construction of a solution. In order to efficiently explore this search tree, BB algorithms maintain an upper bound and estimate lower bounds for partially constructed solutions. Assuming a minimization problem, the upper bound corresponds to the cost of the best solution found so far. During the search process, a lower bound is computed for any partial solution generated, estimating the cost of the best solution that can be constructed by extending it. If this lower bound is greater than the current upper bound, solutions constructed by extending it will not lead to an improvement, and thus all nodes descending from it can be pruned from the search tree. Clearly, the capability of the algorithm for pruning the search tree depends on the existence of an accurate lower bound, which should also be computationally inexpensive in order to be practical.

Beam search (BS) (Barr & Feigenbaum, 1981) algorithms are incomplete derivates of BB algorithms, and are thus heuristic methods. Essentially, BS works by extending every partial solution from a set B (called the *beam*) in at most $k_{ext}$ possible ways. Each new partial solution so generated is stored in a set B'. When all solutions in B have been processed, the algorithm constructs a new beam by selecting the best up to $k_{bw}$ (called the *beam width*) solutions from B'. Clearly, a way of estimating the quality of partial solutions, such as a lower bound, is needed for this.

An interesting peculiarity of BS is the way it extends in parallel a set of different partial solutions in several possible ways, making it a particularly suitable tree search method to be used in a hybrid collaborative framework (it can be used to provide periodically promising partial solutions to a population-based search method such as a memetic algorithm). Gallardo, Cotta, and Fernández (2007) have shown that this kind of hybrid algorithms can provide excellent results for some combinatorial optimization problems. We will subsequently present a hybrid tree search/memetic algorithm for WCSPs based on this idea.

## 2.5 Memetic Algorithms

Evolutionary algorithms (EAs) are population-based metaheuristic optimization methods inspired by biological evolution (Bäck et al., 1997). In order to explore the search space, the EA maintains a set of solutions known as the *population* of *individuals*. These are usually randomly initialized across the search space, although heuristics may also be used. After the initialization, three different phases are iteratively performed until a termination condition is reached: *selection*, *reproduction* (which encompasses *recombination* and *mutation*) and *replacement*. In the context of EAs, the objective function assigning values to each solution is termed a *fitness function*, and is used to guide the search.

Note that EAs are black box optimization procedures in the sense that no knowledge of the problem (apart from the fitness function) is used. The need to exploit problem-knowledge has been repeatedly shown (Wolpert & Macready, 1997; Culberson, 1998) however. Different attempts have been made to answer this need; Memetic algorithms (Moscato & Cotta, 2003, 2007; Krasnogor & Smith, 2005) (MAs) are one of the most successful approaches to date (Hart et al., 2005). Like EAs, MAs are also population based metaheuristics. The main difference is that the components of the population (sometimes termed *agents* in MA terminology) are not passive entities. Rather, they are active entities that cooperate and compete in order to find improved solutions.

There are many possible ways to implement MAs. The most common implementation consists of combining an EA with a procedure to perform a local search on some or all solutions in the population during the main generation loop (cf. Krasnogor & Smith, 2005). Figure 2 shows the general outline of such a MA; $p_X$, $p_m$ and *arity* respectively refer to the recombination probability, mutation probability and recombination arity – i.e., number of parents involved in recombination. It must be noted however that the MA paradigm does not simply reduce itself to this particular scheme and there are different places (e.g., population initialization, genotype to phenotype mapping, evolutionary operators, etc.) where problem specific knowledge can be incorporated. In this work, in addition to using tabu search (Glover, 1989, 1990) (TS) as a local search procedure within the MA, we have designed an "intelligent" recombination operator that uses a relaxation of bucket elimination in order to find the best solution that can be constructed from a set of parents without introducing implicit mutation (i.e., exogenous information).

---

**Memetic Algorithm**

---

        **function** MA $(p_X, p_m, arity)$
1:   **for** $i := 1$ **to** $popsize$ **do**
2:      $pop[i] :=$ Random solution$(n)$
3:      $pop[i] :=$ Local Search$(pop[i])$
4:      Evaluate$(pop[i])$
5:   **end for**
6:   **while** no timeout **do**
7:      **for** $i := 1$ **to** $offsize$ **do**
8:        **if** recombination is performed (under $p_X$) **then**
9:          **for** $j := 1$ **to** $arity$ **do**
10:            $parent_j :=$ Select$(pop)$
11:          **end for**
12:          $offspring[i] :=$ Recombine$(parent_1, parent_2, \dots, parent_{arity})$
13:        **else**
14:          $offspring[i] :=$ Select$(pop)$
15:        **end if**
16:        **if** mutation is performed (under $p_m$) **then**
17:          $offspring[i] :=$ Mutate$(offspring[i])$
18:        **end if**
19:        $offspring[i] :=$ Local Search$(offspring[i])$
20:        Evaluate$(offspring[i])$
21:      **end for**
22:      $pop :=$ Replace$(pop, offspring)$
23:   **end while**

---

Figure 2: Pseudo code of a memetic algorithm (MA). Although different variants are possible with respect to this scheme, it broadly captures a typical algorithmic structure of MAs.

## 3. A Multi-Level Memetic/Exact Hybrid Algorithm for WCSPs

WCSPs are very suitable for being tackled with evolutionary metaheuristics. Obviously, the quality of the results will greatly depend on how well knowledge of the problem is incorporated into the search mechanism. Our final goal is to present an algorithmic model based on the hybridization of MAs with exact techniques at two levels: within the MA (as an embedded operator), and outside it (in a cooperative model). Firstly, we will focus in the next subsection on the first level of hybridization, which incorporates an exact technique (namely BE) within the MA as an embedded recombination operator. Subsequently, we will proceed to a second level of hybridization, in which the MA cooperates with a branch-and-bound based beam search algorithm that further uses the technique of mini-buckets as a lower bound (see Figure 3).

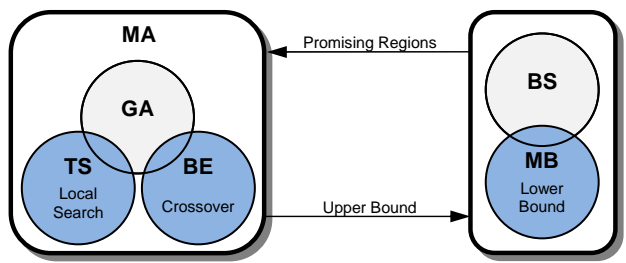

Figure 3: Schematic description of the proposed hybrid algorithm.

## 3.1 Optimal recombination with BE

As previously mentioned, one of the phases that constitutes a typical MA is recombination (i.e., lines 9-14 in Figure 2), in which some individuals in the population are combined with the aim of obtaining improved individuals. For this purpose, different standard recombination operators have been proposed in the literature (see Bäck et al., 1997). Although those blind operators are feasible from a computational point of view, they would perform poorly, as no problem knowledge is being used. In the context of WCSPs, we can resort to BE in order to achieve a sensible recombination of information.

Even though the performance of BE as an exact method for the resolution of WCSPs may be better than basic search-based approaches, the corresponding time and space complexity can still be very high, making this technique unsuitable for large instances. In the following, we explain how BE can be used to implement an intelligent recombination operator for WCSPs. Such operator will implicitly explore the possible children of the solutions being recombined, providing the best solution that can be constructed without introducing implicit mutation, i.e., exogenous information (cf. Cotta & Troya, 2003). Note that this use of bucket elimination is related to what is usually referred to as Large Neighborhood Search (Ahuja, Ergun, Orlin, & Punnen, 2002).

For the sake of simplicity, let us assume that all variables in WCSP $(\mathcal{X}, \mathcal{D}, F)$ have the same domain (i.e., $D_1 = \cdots = D_n$), and let $x = (x_1, x_2, \cdots, x_n)$ and $y = (y_1, y_2, \cdots, y_n)$ be two solutions to be recombined, and $[z_i]$ will be the value of variable $z_i$. Our operator will calculate the best solution that can be obtained by combining variables from $x$ and $y$ without introducing information not present in any of the parents. This can be achieved by restricting the domain of variables in BE to values appearing in the configurations being recombined. The recombination operator becomes BE$(\mathcal{X}, D, \mathcal{F})$, where $D = \{[x_1], \cdots, [x_n], [y_1], \cdots, [y_n]\}$. Applying this approach to a WCSP in which variables may have different domains would require previously separating the set of variables $\mathcal{X}$ into subsets of variables sharing the same domain.

## 3.2 A Beam Search/MA Hybrid Algorithm

In this subsection, we describe a hybrid tree search/memetic algorithm for WCSPs. This algorithm combines, in a collaborative way (Puchinger & Raidl, 2005), a BS algorithm and a MA. As noted previously, BS works by extending in parallel a set of different partial solutions in several possible ways, and thus can be used to provide promising partial solutions to a

---

**Hybrid algorithm for a WCSP**

---

**function** BS-MA $(\mathcal{X}, \mathcal{D}, k_{bw}, k_{MA})$

1:    $sol := \infty$
2:    $\mathcal{B} := \{ \, () \, \}$
3:    **for** $i := 1$ **to** $n$ **do**
4:        $\mathcal{B}' := \{\}$
5:        **for** $s \in \mathcal{B}$ **do**
6:           **for** $a \in D_i$ **do**
7:              $\mathcal{B}' := \mathcal{B}' \cup \{s \cdot (x_i = a)\}$
8:           **end for**
9:        **end for**
10:      $\mathcal{B} := \textbf{select}$ best $k_{bw}$ nodes from $\mathcal{B}'$
11:      **if** $(i \geqslant k_{MA})$ **then**
12:         **initialize** MA population with best *popsize* nodes from $\mathcal{B}'$
13:         **run** MA
14:         $sol := \textbf{min}\,(sol, \text{MA solution})$
15:      **end if**
16:    **end for**
17:    **return** $sol$
**end function**

---

Figure 4: Hybrid algorithm for a WCSP.

population based search method such as a MA. The goal is to exploit the capability of BS for identifying potentially good regions of the search space, and also to exploit the MA to explore these regions, synergistically combining these two different approaches.

The proposed hybrid algorithm, that executes BS and the MA in an interleaved way, is depicted in Figure 4. In the pseudo-code, a (possibly partial) solution for a WCSP instance is represented by a vector of variables $s = (x_1, x_2, \ldots, x_i)$, $i \leqslant n$, where $s \cdot (x_i = a)$ stands for the extension of partial solution $s$ by assigning value $a$ to its $i$-th variable as noted previously. The hybrid algorithm constructs a search tree, such that its leaves consist of complete solutions and internal nodes at level $i$ represent partially specified (up to the $i$-th variable) solutions. This tree is heuristically traversed in a breadth first way using a BS algorithm with beam width $k_{bw}$ (i.e., maintaining only the best $k_{bw}$ nodes at each level of the tree). For the beam selection (line 10), a heuristic quality measure has to be defined for partial solutions, whose value must be $\infty$ if the partial solution is unfeasible. The algorithm starts (line 2) with a totally unspecified solution. Initially, only the BS part of the algorithm is executed. During each iteration of BS (lines 3-17), a new variable is assigned for every solution in the beam (line 7). The interleaved execution of the MA starts only when partial solutions in the beam have at least $k_{MA}$ variables (line 11). For each iteration of BS, the best *popsize* solutions in the beam are selected (using the quality measure described above) to initialize the population of the MA (line 12). Since these are partial solutions, they must be first converted into full solutions, e.g., by completing remaining variables randomly.

After running the MA, its solution is used to update the incumbent solution (*sol*), and this process is repeated until the search tree is exhausted.

### 3.3 Computing Tight Bounds with Mini-Buckets

The performance of the BS component of the algorithm described in the previous section will depend on the quality of the heuristic function used to estimate partial solutions (line 10 of Figure 4). In order to compute a tight, yet computationally inexpensive, lower bound for the remanning part of the solution we can resort to Mini-Buckets (MB). As described by Kask and Detcher (2001), the intermediate functions created by applying the MB scheme can be used as a general mechanism to compute heuristic functions that estimate the best cost of yet unassigned variables in partial solutions. To this end, MB must be run as a preprocessing stage, using the reverse order in which the search will instantiate variables. The set of augmented buckets computed during this process can be used as estimations of the best cost extension to partial solutions (check the work of Kask & Detcher, 2001, for details).

## 4. Tackling the Maximum Density Still Life Problem

Previously proposed algorithms are general enough to be used in many WCSPs in which BE can be executed. In this section we present an application case study on the maximum density still life problem (MDSLP). This problem is defined in the context of the game of life proposed by John H. Conway in the 60s and divulged by Martin Gardner (Gardner, 1970), so let us first describe this game. It is played on an infinite checkerboard in which the only player places checkers on some of its squares. Each square on the board is called a cell and has eight neighbors; the eight cells that share one or two corners with it. A cell is alive if there is a checker on it, and is dead otherwise. The contents of the board evolve iteratively, in such a way that the state at time $t$ determines the state at time $t + 1$ according to some simple rules: (1) a live cell remains alive if it has two or three live neighbors, otherwise it dies, and (2) a dead cell becomes alive it is has exactly three live neighbors.

The simple rules of the game of life can nevertheless generate an incredibly complex dynamics. To better understand the MDSLP, let us define a stable pattern (also called a *still life*) as a board configuration that does not change over time, and let the *density* of a region be its percentage of living cells. The MDSLP in an $n \times n$ grid consists of finding a *still life* of maximum density. Elkies (1998) has shown that, for infinite boards, the maximum density is $1/2$ (for finite size, no exact formula is known). In this paper, we are concerned with the MDSLP and finite patterns, that is, finding maximal $n \times n$ still lifes.

### 4.1 Related Work

The MDSLP has been tackled in the literature using different approaches. Bosch and Trick (2002) compared different formulations for the MDSLP using integer programming (IP) and constraint programming (CP). Their best results were obtained with a hybrid algorithm mixing the two approaches. They were able to solve the cases for $n = 14$ and $n = 15$ in about 6 and 8 days of CPU time respectively. Smith (2002) used a pure constraint programming approach to address the problem. However, only instances up to $n = 10$ could be solved. The best results for this problem were reported by Larrosa and Morancho (2003) and Larrosa et al. (2005), showing the usefulness of *bucket elimination* (BE), an exact technique based on variable elimination and commonly used for solving constraint satisfaction problems described in detail in Section 2.2. Their basic approach could solve the problem for $n = 14$ in about $10^5$ seconds. Further improvements increased the boundary to $n = 20$ in about twice as much time. Recently, Cheng and Yap (2005, 2006) have tackled the problem via the use of ad-hoc global `case` constraints, but their results are comparable to IP/CP hybrids, and thus cannot be compared to the ones obtained previously by Larrosa et al.

Table 1: Best experimental results reported by Bosch and Trick (2002) (CP/IP), Larrosa and Morancho (2003) (BE) and Larrosa et al. (2005) (HYB-BE) for solving the MDSLP. Time is indicated in seconds.

| | 12 | 13 | 14 | 15 | 16 | 17 | 18 | 19 | 20 |
|---|---|---|---|---|---|---|---|---|---|
| optimum | 68 | 79 | 92 | 106 | 120 | 137 | 153 | 171 | 190 |
| CP/IP | 11536 | 12050 | $5 \times 10^5$ | $7 \times 10^5$ | | | | | |
| BE | 1638 | 13788 | $10^5$ | | | | | | |
| HYB-BE | 1 | 2 | 2 | 58 | 7 | 1091 | 2029 | 56027 | $2 \times 10^5$ |

Table 1 summarizes experimental results for current approaches used to tackle the MD-SLP, reporting the computational times of the hybrid IP/CP algorithm of Bosch and Trick (2002), the BE approach of Larrosa and Morancho (2003) and the BE/search hybrid of Larrosa et al. (2005). Although different computational platforms may have been used for these experiments, the trends are very clear and give a clear indication of the potential of the different approaches. It should be noted that all these techniques applied to the MDSLP are exact approaches. They are inherently limited for increasing problem sizes and their capabilities as anytime algorithms are unclear. To tackle this problem, we recently proposed the use of hybrid methods combining exact and metaheuristic approaches. We considered the hybridization of BE with evolutionary algorithms (a stochastic population-based search method) endowed with tabu search (a local search method)(Gallardo, Cotta, & Fernández, 2006a). The resulting algorithm was a memetic algorithm (MA; see Section 2.5). It used BE as a mechanism for recombining solutions, providing the best possible child from the parental set. Experimental tests indicated that the algorithm provided optimal or near-optimal results at an acceptable computational cost. Subsequently, we studied extended multi-level models in which our previous hybrid algorithm was further hybridized with a branch-and-bound derivative, namely beam search (BS)(Gallardo, Cotta, & Fernández, 2006b). Studies on the influence that variable clustering and multi-parent recombination have on the performance of the algorithm were also conducted. The results indicated that variable clustering was detrimental for this problem but also that multi-parent recombination improves the performance of the algorithm. To the best of our knowledge, these are the only heuristic approaches applied to this problem to date.

In this section, our previous research on this problem is included and extended. As new contributions, we have redone all the experiments using an improved implementation of the bucket elimination crossover operator, described in Section 3.1. Additionally, we present a more extensive experimental analysis of our BS/MA hybrid described in (Gallardo et al., 2006b), analyzing the sensitivity of its parameters. We also propose a new hybrid algorithm that uses the technique of mini-buckets to further improve the lower bounds of the partial solutions considered in the BS part of the hybrid algorithm. This new algorithm is obtained from the hybridization, at different levels, of complete solving techniques (BE), incomplete deterministic methods (BS and MB) and stochastic algorithms (MAs). An experimental analysis shows that this new proposal consistently finds optimal solutions for MDSLP instances up to $n = 20$ in considerably less time than all the previous approaches reported in the literature. Finally, in order to test the scalability of our approach, this novel hybrid algorithm has been run on very large instances of the MDSLP for which an optimal solution is currently unknown. The results were very successful, as the algorithm performed at the state-of-the-art level, providing solutions that are equal or better than the best ones reported to date in the literature. For readability reasons, many particular technical details of the different algorithms for the MDSLP are omitted, but are fully described in an accompanying report (Gallardo, Cotta, & Fernández, 2008). At any rate, a model of the MDSLP as a WCSP is presented in Appendix A.

## 4.2 A Memetic Algorithm for the MDSLP

First of all, we develop a MA for the MDSLP. In this MA, an $n \times n$ board is represented as a binary $n \times n$ matrix. Based on the stratified gradient provided by a penalty based fitness function that measures the number of violated constraints and their distance to feasibility (prioritizing the former over the latter), an efficient local search strategy that explores the set of solutions obtained by flipping exactly one cell in a configuration was devised. In order to escape from local optima, a tabu-search scheme is used (line 19 in Figure 2).

The MA uses BE as a crossover operation as described in Section 3.1 (line 12 in Figure 2). One interesting property of the operator described is that it is not limited to recombining only two board configurations, but can instead be generalized to recombine any number of them by considering domains consisting of all the values a variable has in any of the parents. This multi-parental capability has also been explored in the rest of this paper.

To evaluate the usefulness of the described hybrid recombination operator, a set of experiments for problem sizes from $n = 12$ up to $n = 20$ has been realized (recall that optimal solutions to the MDSLP are known up to $n = 20$). The experiments have been performed using a steady-state evolutionary algorithm ($popsize = 100$, $p_m = 1/n^2$, $p_X = 0.9$, binary tournament selection). With the aim of maintaining diversity, duplicated individuals are not allowed in the population. Algorithms were run until an optimal solution was found or a time limit exceeded. This time limit was set to 3 minutes for problem instances of size 12 and was gradually increased by 60 seconds for each size increment. For each algorithm and each instance size, 20 independent runs have been made. All experiments in this paper have been performed on a Pentium IV PC (2400MHz and 512MB RAM) under SuSE Linux.

The base algorithm is a MA using a two-dimensional version of SPX (single-point crossover) for recombination, and endowed with tabu search for local improvement. This algorithm is termed MA$_{TS}$, and has been shown to be capable of finding feasible solutions systematically, solving to optimality instances with $n < 15$ (see MA$_{TS}$ in Figure 5a).

Although the performance of the algorithm degrades for larger instances, it provides distributions for the solutions whose average relative distance to the optimum is less than 5.29% in all cases. This contrasts with the case of plain EAs, which are incapable of finding even a feasible solution in most runs (Gallardo et al., 2006a).

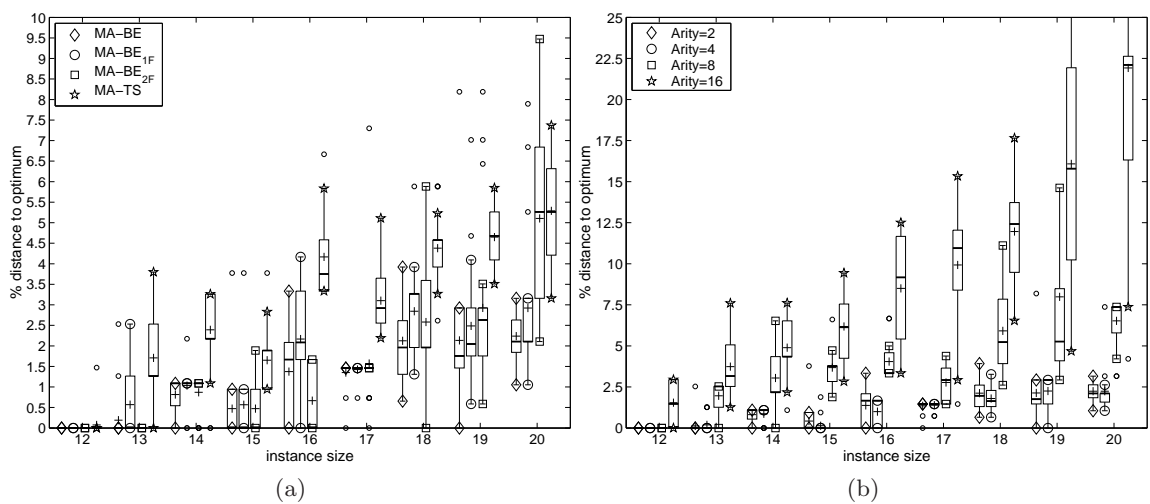

(a)                                    (b)

Figure 5: Relative distances to optimum for different (a) algorithms and (b) arities for sizes ranging from 12 up to 20. In this and in all subsequent figures, each box summarizes 20 runs, boxes comprise the second and third quartiles of the distribution (i.e., the inner 50%), a horizontal line marks the median, a plus sign indicates the mean, and circles indicate results further from the median than 1.5 times the interquartile-distance.

MA$_{TS}$ is firstly compared with MAs endowed with BE for performing recombination. Since the use of BE for recombination has a higher computational cost than a simple blind recombination, and there is no guarantee that recombining two infeasible solutions will result in a feasible solution, we have defined three variants of the MAs:

- In the first one, called MA-BE, BE is always used to perform recombination.

- In the second, termed MA-BE$_{1F}$, we require that at least one of the parents be feasible in order to apply BE; otherwise blind recombination is used.

- In the last variant, identified as MA-BE$_{2F}$, we require the two parents to be feasible, thus being more restrictive in the application of BE.

By evaluating these variants, we intend to explore the computational tradeoffs involved in the application of BE as an embedded component of the MA. For these algorithms, mutation is performed prior to recombination in order to take advantage of good solutions provided by BE. Figure 5a shows the empirical performance of the different algorithms. Results show that MA-BE returns significantly better results than MA$_{TS}$. MA-BE$_{2F}$ can find slightly better solutions than MA-BE on smaller instances ($n \in \{13, 15, 16\}$), but on larger instances the winner is MA-BE. It seems that the effort saved by not recombining unfeasible solutions does not further improve the performance of the algorithm. Note also that, for larger instances, MA-BE$_{1F}$ is better than MA-BE$_{2F}$. This correlates well with the fact that BE is used more frequently in the former than in the latter.

As mentioned in Section 3.1, the optimal recombination scheme we use can be readily extended to multi-parent recombination (Eiben, Raue, & Ruttkay, 1994): an arbitrary number of solutions can contribute their constituent rows for constructing a new solution. Additional experiments were done to explore the effect of this capability of MA-BE. Figure 5b shows the results obtained by MA-BE for a different number of parents being recombined (arities 2, 4, 8 and 16). For $arity = 2$, the algorithm was able to find the optimum solution for all instances except for $n = 18$ and $n = 20$ (the relative distance to the optimum for the best solution found is less than 1.04% in these cases). Runs with $arity = 4$ cannot find optimum solutions for the remaining instances, but note that the distribution improves in some cases. Clearly, the performance of the algorithm deteriorates when combining more than 4 parents due to the higher computational cost of BE. Variable clustering could be used to alleviate this higher computational cost, but this results in performance degradation since the coarser granularity of the pieces of information hinders information mixing (Cotta & Troya, 2000; Gallardo et al., 2006b).

### 4.3 A BS/MA Hybrid Algorithm for the MDSLP

In this section we evaluate an instantiation of the BS and MA hybrid algorithm described in Section 3.2 for the MDSLP, called BS-MA-BE. For the beam selection (line 10 in Figure 4), a simple quality measure is defined for partial solutions, whose value is either $\infty$ if the partial configuration is unstable, or its number of dead cells otherwise. The methodology is the same as in Section 4.2 (20 executions are performed for each algorithm and instance size), but arities for the MA are in $\{2, 3, 4\}$. The setting of the remaining parameters is $k_{bw} = 2000$ (preliminary tests indicated that this value was reasonable), and $k_{MA} \in \{0.3 \cdot n, 0.5 \cdot n, 0.75 \cdot n\}$, i.e., the best 2000 nodes were kept on each level of the BS algorithm, and 30%, 50% or 75% of the levels of the BS tree were initially descended before the MA was run. With respect to termination conditions, each execution of the MA within the hybrid algorithm consists of 1000 generations, and no time limits are imposed for the hybrid algorithms, which are run for $n$ iterations of the BS.

Figure 6a shows the results for different values of parameter $k_{MA}$. In order to better compare the distributions, the number of optimal solutions obtained by each algorithm (out of 20 executions) is shown above each box plot. For $k_{MA} = 0.3 \cdot n$, the performance of the resulting algorithm improves significantly over the original MA. Note that BS-MA-BE, using an arity of 2 parents, is able to find the optimum for all cases except for $n = 18$ (this instance is solved with $arity = 4$). All distributions for different instance sizes are significantly improved. For $n < 17$ and $arity \in \{2, 3, 4\}$, the algorithm consistently finds the optimum in all runs. For other instances, the solution provided by the algorithm is always within 1.05% of the optimum, except for $n = 18$, for which the relative distance to the optimum for the worst solution is 1.3%. The other two charts show that, in general, the performance of the algorithm deteriorates with increasing values of the $k_{MA}$ parameter. This may be due to the low quality of the bounds used in the BS part.

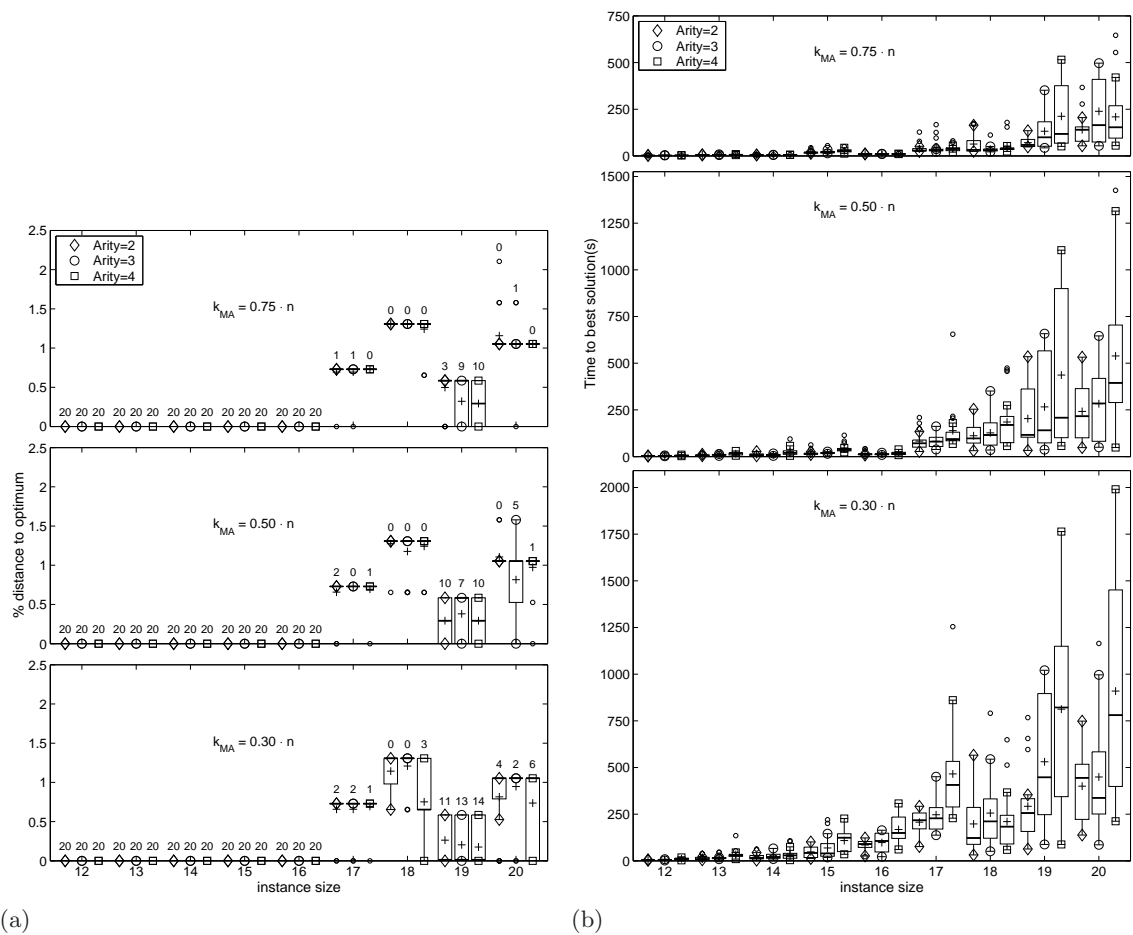

(a)

(b)

Figure 6: (a) Relative distances to optimum and (b) time to best solution for different arities for BS-MA-BE and $K_{MA} \in \{0.3 \cdot n, 0.5 \cdot n, 0.75 \cdot n\}$, for sizes ranging from 12 up to 20. The numbers above each box indicate how many times the optimal solution was found.

Regarding execution times, Figure 6b shows time distributions (in seconds) to reach the best solution needed by the algorithms. Although BS-MA-BE requires more time than MA-BE, the time needed remains reasonable for these instances, and is always less than 2000 seconds. Note also how the execution time increases with the arity, as more time is needed by the MA to perform BE in the crossover operator. On the other hand, execution time decreases for larger values of $k_{MA}$ as the number of executions of the MA decreases, although, as we have already remarked, the quality of the solutions worsens.

## 4.4 Improving the Lower Bound using MB for the MDSLP

The simple quality measure for beam selection used in the previous section depends solely on the part of the solution that is already constructed. In this section, we will experimentally study the use of the MB technique to compute a tight, yet computationally inexpensive, lower bound for the remanning part of the configuration with the aim of improving the performance of the BS part of the hybrid algorithm. Basically, the idea is to cluster all cells in the same row of the board in a metavariable. These metavariables can be partitioned into $M$ columns with $\approx n/M$ cells each. Finally, we can resort to MB to estimate best cost extensions to a partial board configuration by considering only each of the columns. By summing estimations for all column extensions, a bound on the best board extension to a partial solution is obtained. In this section, we have experimented with $M = 3$ (i.e., three columns for each row), although if the complexity is still too high, the same approach can be used to reduce it further, by considering more columns.

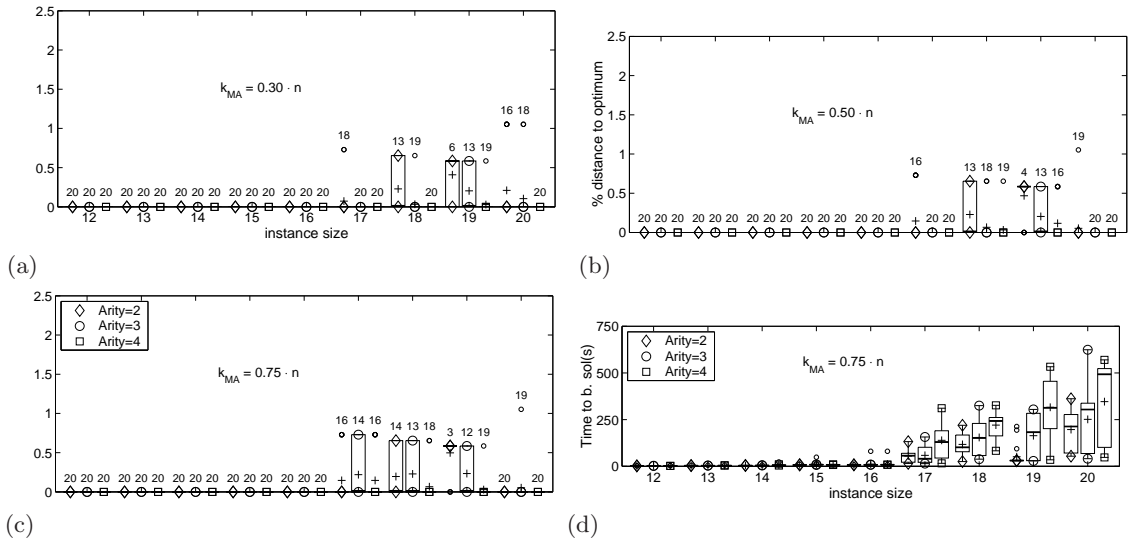

Figure 7: (a)-(c) Relative distances to the optimum using different arities for BS-MA-BE-MB and $K_{MA} \in \{0.3 \cdot n, 0.5 \cdot n, 0.75 \cdot n\}$, for sizes ranging from 12 up to 20. (d) Time (in seconds) to best solution for different arities for BS-MA-BE-MB and $k_{MA} = 0.75 \cdot n$, for sizes ranging from 12 up to 20.

Experiments were repeated for the hybrid algorithm equipped with the new lower bound, BS-MA-BE-MB. Figure 7a-7c shows the results of these experiments for values of $k_{MA} \in \{0.3 \cdot n, 0.5 \cdot n, 0.75 \cdot n\}$. The algorithm finds the optimum for all instances and arities and the relative distance to the optimum for the worst solution found is less than 1.05% in all cases. The best results are obtained with $arity = 4$, although this requires slightly more execution time. Note also how BS-MA-BE-MB is less sensitive to the setting of parameter $k_{MA}$, which means that execution times can be reduced considerably using a large value for this parameter (see Figure 7d). The particular combination of parameters $k_{MA} = 0.75 \cdot n$ and $arity = 4$ provides excellent results at a lower computational cost, as execution times are always below 570 seconds for $n \leqslant 20$. As a comparison, recall that the only approach in the literature that can solve these instances – described by Larrosa et al. (2005) – requires over 33 minutes for $n = 18$, 15 hours for $n = 19$ and 2 days for $n = 20$, and that other approaches are unaffordable for $n > 15$. Note however that these times correspond to a computational platform different to ours. In order to make a fairer comparison, we executed the algorithm of Larrosa et al. [3] in our platform. In this case, it required 1867 seconds (i.e., more than 31 minutes) in order to solve the $n = 18$ instance, and more than 1 day and 18 hours to solve the $n = 20$ instance. These values are very close to the times reported by Larrosa et al. (2005), and hence indicate that the computational platforms are fairly comparable.

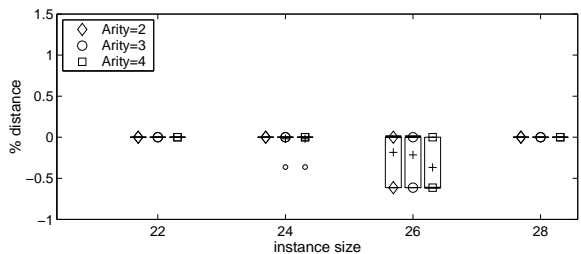

Figure 8: Relative distances to best known solutions using different arities for BS-MA-BE-MB and $k_{MA} = 0.3 \cdot n$, for very large instances (i.e., sizes of 22, 24, 26, and 28). Note the improvement of best known solutions for sizes 24 and 26.

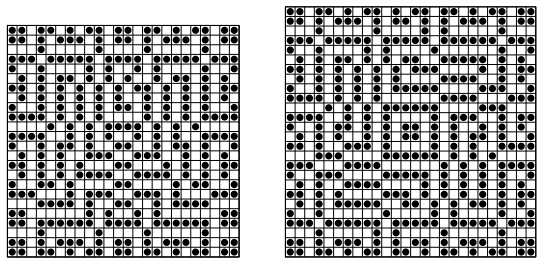

Figure 9: New best known maximum density still lifes for $n \in \{24, 26\}$.

Table 2: Optimal solutions for the SMDLP.

| $n$ | 12 | 13 | 14 | 15 | 16 | 17 | 18 | 19 | 20 | 22 | 24 | 26 | 28 |
|-----|----|----|----|-----|-----|-----|-----|-----|-----|-----|-----|-----|-----|
| opt | 68 | 79 | 92 | 106 | 120 | 137 | 154 | 172 | 192 | 232 | 276 | 326 | 378 |

## 4.5 Results on Very Large Instances

As already mentioned, there is currently no approach available to tackle the MDSLP for $n > 20$. Larrosa et al. (2005) tried their algorithm for $n = 21$ and $n = 22$, but they could not solve any of those instances within a week of CPU. For these very large instances, only solutions to some relaxations of the problem are known. One of these relaxations, known as the symmetrical maximum density still life problem (SMDSLP), was proposed by Bosch and Trick (2002), and consists of considering only symmetric boards (either horizontally or vertically) which reduces the search space from $2^{n^2}$ to $2^{n\lceil n/2 \rceil}$.

---

3. Available at `http://www.lsi.upc.edu/~larrosa/publications/LIFE-SOURCE-CODE.tar.gz` . Time for $n = 19$ could not be obtained as the code provided by Larrosa et al. can only be used with even sized instances.

BE alone can find vertically symmetric still lifes, by considering as variable domains sets that contain only symmetric rows. Larrosa and Morancho (2003) and Larrosa et al. (2005) used this algorithm to solve the SMDSLP for the instances considered so far in this paper (i.e., for $n \in \{12 .. 20\}$), as well as for very large instances (i.e., $n \in \{22, 24, 26, 28\}$). The results are summarized in Table 2, which shows for each instance size the optimal symmetrical solution (as the number of dead cells). Clearly, the cost of optimal symmetric still lifes are upper bounds for the MDSLP, which can additionally be observed to be very tight for $n \leqslant 20$. Results for $n > 20$ are currently the best known solutions for these instances.

We also run our algorithm (BS-MA-BE-MB) for these very large instances (i.e., $n \in \{22, 24, 26, 28\}$), and compare our results to symmetrical solutions for these instances. Results (displayed in Figure 8) show that our algorithm is able to find two new best known solutions for the MDSLP, namely for $n = 24$ and $n = 26$. There are 275 and 324 dead cells respectively in the new solutions. These solutions are pictured in Figure 9. It is also worth noting that our algorithm could also find a solution with 325 dead cells for the $n = 26$ instance. For the other instances, our algorithm could reach the best known solutions consistently. The computation of mini-Buckets for these very large instances is done by considering four clustered cost functions for variables in each row of the board, as the complexity when using three cost functions was still too high.

## 5. Conclusions

Many problems can be modeled as WCSPs. One exact technique that has been used to tackle such problems is BE. However, the high space complexity of BE as an exact technique, makes this approach impractical for large instances. In this case, one can resort to mini-buckets to get an approximate solution, although the complexity can again be large. In this work, we have presented several proposals for the hybridization of BE and MB with memetic algorithms and beam search in order to get effective heuristics and have shown that they represent very promising models.

We have experimentally evaluated our model with the MDSLP, an excellent example of WCSP. Its highly constrained nature is typical in many optimization scenarios. The difficulty of solving this problem illustrates the limitations of classical optimization approaches, and highlights the capabilities of the proposed approaches. Indeed, the experimental results have been very positive, solving large instances of the MDSLP to optimality. Among the different models presented, we must distinguish a new algorithm resulting from the hybridization, at different levels, of complete solving techniques (i.e., bucket elimination), incomplete deterministic methods (i.e., beam search and mini-buckets) and stochastic algorithms (i.e., memetic algorithms). This algorithm empirically produces good-quality results, not only solving to optimality very large instances of the constrained problem in a relatively short time, but also providing new best known solutions in some large instances.

As future work, we plan to consider complete versions of the hybrid algorithm. This involves the use of appropriate data structures to store not yet considered but promising branch-and-bound nodes. While the memory requirements will of course grow enormously with the size of the problem instance considered, it will be interesting to analyze the computational tradeoffs of the algorithm as an anytime technique.

## Acknowledgments

We would like to thank Javier Larrosa for his valuable comments, which helped us to improve significantly a preliminary version of this paper. Thanks are also due to the reviewers for their constructive comments. This work has been partially supported by Spanish MICInn under grant TIN2008-05941 (Nemesis).

## Appendix A. The MDSLP as a WCSP

As shown by Larrosa and Morancho (2003) and Larrosa et al. (2005), the MDSLP can be well formulated as a WCSPs. To this end, an $n \times n$ board configuration can be represented by an $n$-dimensional vector $(r_1, r_2, \ldots, r_n)$. Each vector component encodes (as a binary string) a row, so that the $j$-th bit of row $r_i$ (noted $r_{ij}$) represents the state of the $j$-th cell of the $i$-th row (a value of 1 represents a live cell and a value of 0 a dead cell).

Two functions over rows will be useful to describe the constraints that must be satisfied by a valid configuration. The first one,

$$zeroes(a) = \sum_{1 \leqslant i \leqslant n} (1 - a_i), \tag{1}$$

returns the number of dead cells in a row (i.e., the number of zeroes in binary string $a$). The second one,

$$Adjs(a) = Adjs'(a, 1, 0) \tag{2}$$

$$Adjs'(a, i, l) = \begin{cases} l, & i > n \\ Adjs'(a, i+1, l+1), & a_i = 1 \\ \max(l, Adjs'(a, i+1, 0)), & a_i = 0, \end{cases}$$

computes the maximum number of adjacent living cells in row $a$. We also introduce a ternary predicate, $Stable(r_{i-1}, r, r_{i+1})$, which takes three consecutive rows in a board configuration and is satisfied if, and only if, all cells in the central row are stable (i.e., all cells in row $r$ will remain unchanged in the next iteration of the game):

$$Stable(a, b, c) = \bigwedge_{1 \leqslant i \leqslant n} S(a, b, c, i) \tag{3}$$

$$S(a, b, c, i) = \begin{cases} 2 \leqslant \eta(a, b, c, i) \leqslant 3, & b_i = 1 \\ \eta(a, b, c, i) \neq 3, & b_i = 0 \end{cases}$$

$$\eta(a, b, c, i) = \sum_{\max(1, i-1) \leqslant j \leqslant \min(n, i+1)} (a_j + b_j + c_j) - b_i,$$

where $\eta(a, b, c, i)$ is the number of living neighbors of cell $b_i$, assuming $a$ and $c$ are the rows above and below row $b$.

The MDSLP can now be formulated as a WCSP using $n$ cost functions $f_i$, $i \in \{1 \mathrel{..} n\}$. Accordingly, $f_n$ is binary with scope the last two rows of the board ($var(f_n) = \{r_{n-1}, r_n\}$) and is defined as:

$$f_n(a, b) = \begin{cases} \infty, & \neg Stable(a, b, \theta) \vee Adjs(b) > 2 \\ zeroes(b), & \text{otherwise.} \end{cases} \qquad (4)$$

The first line checks that all cells in row $r_n$ are stable, whereas the second one checks that no new cells are produced below the $n \times n$ board. Note that any pair of rows representing an unstable configuration is assigned a cost of $\infty$, whereas a stable one is assigned its number of dead cells (to be minimized).

For $i \in \{2 \mathrel{..} n - 1\}$, corresponding $f_i$ cost functions are ternary with scope $var(f_i) = \{r_{i-1}, r_i, r_{i+1}\}$ and are defined as:

$$f_i(a, b, c) = \begin{cases} \infty, & \neg Stable(a, b, c) \vee (a_1 = b_1 = c_1 = 1) \vee (a_n = b_n = c_n = 1) \\ zeroes(b), & \text{otherwise.} \end{cases} \qquad (5)$$

In this case, boundary conditions are checked to the left and right of the board. As regards cost function $f_1$, it is binary with scope the first two rows of the board ($var(f_1) = \{r_1, r_2\}$) and is specified similarly to $f_n$:

$$f_1(b, c) = \begin{cases} \infty, & \neg Stable(\theta, b, c) \vee Adjs(b) > 2 \\ zeroes(b), & \text{otherwise.} \end{cases} \qquad (6)$$